\documentclass[runningheads,a4paper]{llncs}

\usepackage{amssymb}
\usepackage{graphicx}
\usepackage{array}

\usepackage{url}
\urldef{\mailsa}\path|{alfred.hofmann, ursula.barth, ingrid.haas, frank.holzwarth,|
\urldef{\mailsb}\path|anna.kramer, leonie.kunz, christine.reiss, nicole.sator,|
\urldef{\mailsc}\path|erika.siebert-cole, peter.strasser, lncs}@springer.com|

\begin{document}

\mainmatter
\title{VideoSET: Video Summary Evaluation \\ through Text}
\titlerunning{VideoSET: Video Summary Evaluation \\ through Text}
\author{Serena Yeung\and Alireza Fathi\and Li Fei-Fei} 
\authorrunning{VideoSET: Video Summary Evaluation through Text}
\institute{Computer Science Department\\ Stanford University} 
\toctitle{VideoSET: Video Summary Evaluation through Text}
\tocauthor{Yeung et al.}
\maketitle
\begin{abstract}
In this paper we present VideoSET, a method for Video Summary Evaluation
through Text that can evaluate how well a video summary is able to
retain the semantic information contained in its original video. We
observe that semantics is most easily expressed in words, and develop
a text-based approach for the evaluation. Given a video summary, a
text representation of the video summary is first generated, and an
NLP-based metric is then used to measure its semantic distance to
ground-truth text summaries written by humans. We show that our technique
has higher agreement with human judgment than pixel-based distance
metrics. We also release text annotations and ground-truth text summaries
for a number of publicly available video datasets, for use by the
computer vision community.
\end{abstract}
\vspace{-3mm}

\section{Introduction}

In today\textquoteright{}s world, we are surrounded by an overwhelming
amount of video data. The Internet Movie Database (IMDb) contains
over 2.7 million entries, and over 100 hours of video are uploaded
to YouTube every minute. Furthermore, wearable camcorders such as
the GoPro and Google Glass are now able to provide day-long recordings
capturing our every interaction and experience. How can we possibly
hope to consume and browse so much video? 

A key answer to this problem is video summarization. Just as text
summaries have long helped us quickly understand documents and determine
whether to read in more depth, we are now in need of video summaries
to help us browse vast video collections. Imagine searching for wedding
videos on YouTube. It is inefficient to browse through the millions
of results that are returned, but being able to watch a short summary
of each result would make the process tremendously easier. On the
other hand, imagine having hours of video from a GoPro-recorded vacation.
Most people would not want to watch or go through these long recordings,
but a video summary could provide a condensed and viewer-friendly
recap. 

\begin{figure*}
\begin{centering}
\includegraphics[scale=0.45]{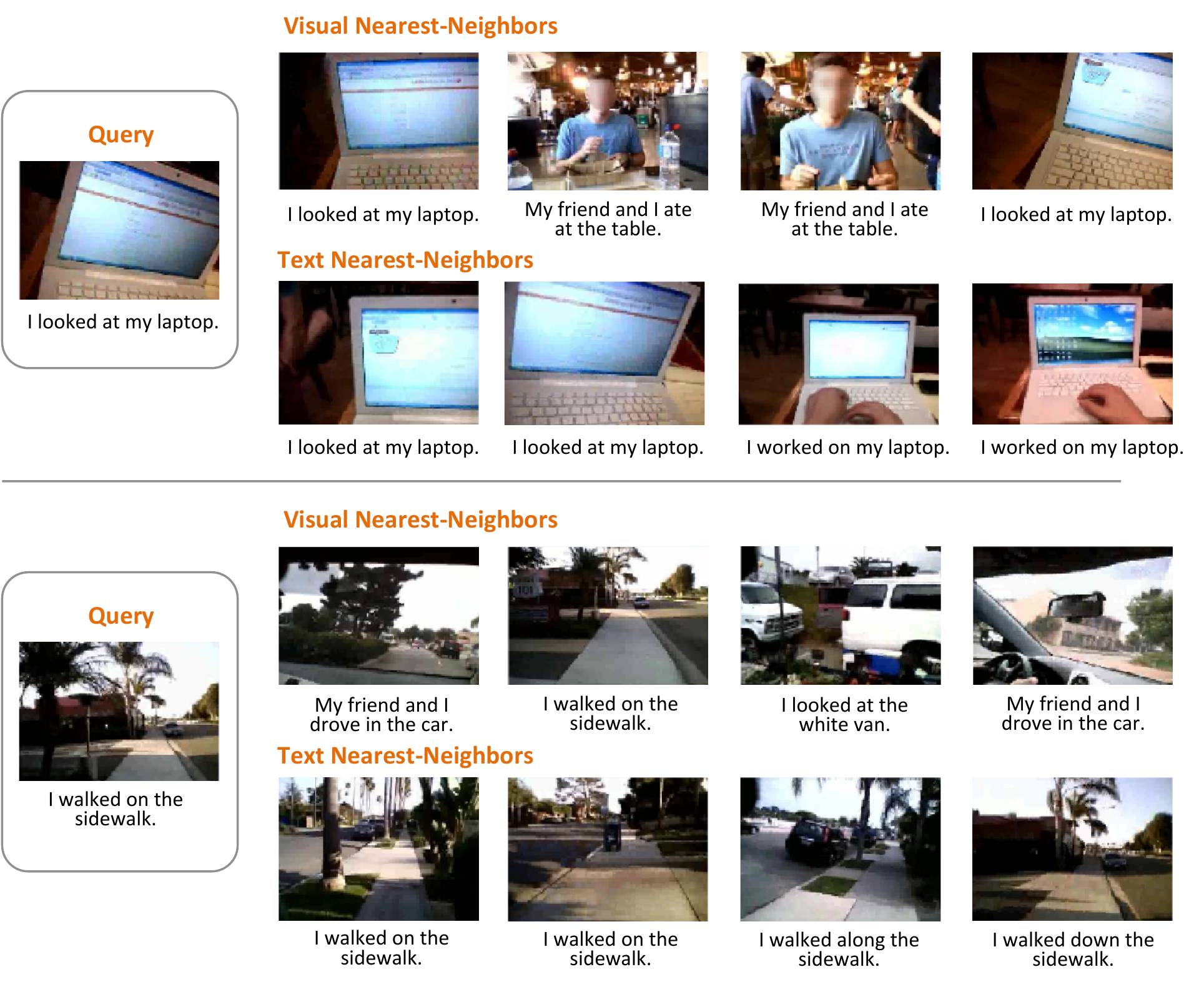}
\par\end{centering}

\caption{Representative images and text descriptions are shown for each query
subshot, and nearest-neighbor subshots based on visual or text distance.
Comparison using text distance is a better indicator of semantic similarity.
\label{fig:Pixel-based-comparison-does}}
\end{figure*}

While the need for video summarization methods is clear, and the computer
vision community has indeed seen a surge of recent interest, development
has been hampered by the lack of a standard, efficient evaluation
method. Most previous work has performed a diverse range of user comparison
studies \cite{Ngo01jig,Doherty08multimedia,Lee12cvpr,Lu13cvpr} that
are difficult to replicate, while a few have used pixel-based comparison
with a ground truth \cite{Li10mm,Khosla13cvpr}. This absence of a
standard can be attributed to a number of challenges. First, how do
we even define what a good summary is? The answer is not obvious,
and user studies have used varied and often vague criteria including
\textquotedblleft{}better overall summary\textquotedblright{}, \textquotedblleft{}better
progress of story\textquotedblright{}, and \textquotedblleft{}representative
of original video\textquotedblright{}. Second, assuming we have a
definition, how do we visually represent an ideal summary, and quantify
the distance of any given summary from this ideal? User comparison
studies try to circumvent this challenge altogether, while pixel-based
comparisons suffer from the problem that visual distance is not an
adequate measure of semantic distance (Fig. \ref{fig:Pixel-based-comparison-does}). 

Our goal in this paper is to address the need for a standard video
summary evaluation framework. We argue that from a user perspective,
an ideal evaluation framework should satisfy the following three properties:
(1) provide a metric that measures the distance of a given summary
from ideal; (2) perform the evaluation in an automated and efficient
manner without human involvement; and (3) provide standard evaluation
datasets on which researchers can compare their summarization methods
against previous work. Due to the challenges discussed above, no evaluation
method to date satisfies these three properties. 

We propose to overcome these challenges using a few key observations.
First, we note that there are indeed many different types of summaries
(e.g. informative substitute, or enticing trailer) that can be defined
and judged in different ways. However, a summary that maximizes semantic
information is extremely useful, and in fact most other types of summaries
can be defined as extensions of this informative summary. Second,
we observe that semantic similarity is most naturally measured through
text. In addition, humans are very good at summarizing information
and experiences in words. As Fig. \ref{fig:Pixel-based-comparison-does}
shows, comparison using the textual descriptions associated with each
image is a much better indicator of semantic similarity. 

Based on these observations, we present VideoSET, a method for Video
Summary Evaluation through Text that can measure how well any summary
retains the \emph{semantic information} of the original video. Given
a video summary to evaluate, our approach first converts the summary
into a text representation, using text annotations of the original
video. It then compares this representation against ground-truth text
summaries written by humans, using Natural Language Processing (NLP)
measures of content similarity. We have obtained and publicly released
all necessary text annotations and ground-truth summaries for a number
of video datasets. In contrast to performing user studies, VideoSET
offers the following important benefits:
\begin{enumerate}
\item It measures the distance of any summary from ideal.
\item It can be easily and reliably replicated.
\item The evaluation is efficient, automated, and requires no human involvement.
\end{enumerate}
In contrast to previous methods using pixel-based comparisons, VideoSET
transfers the evaluation into the text domain to more accurately measure
semantic similarity.

\section{Previous Work}

We group previous work into three sections: (1) methods for video
summarization; (2) techniques for evaluating video summaries; and
(3) techniques for evaluating text summaries. 

\textbf{Methods for video summarization}: Previous methods for video
summarization have used low-level features such as color \cite{Zhang97pr}
and motion \cite{Wolf96assp,Goldman06tog}, or a combination of both
\cite{Dufaux00multimedia}. Some other works have modeled objects
\cite{Kim01multimedia,Liu10pami} and their interaction \cite{Lee12cvpr,Lu13cvpr}
to select key subshots. Kim and Hwang \cite{Kim01multimedia} segment
the objects in video and use the distance between the objects for
video summarization. Liu et al. \cite{Liu10pami} summarize a video
by finding the frames that contain the object of interest. Lee et
al. \cite{Lee12cvpr} find the important objects and people in egocentric
video and select the events that contain them. Lu and Grauman \cite{Lu13cvpr}
model video summarization as a story that relates frames to each other
based on the objects they contain. Khosla et al. \cite{Khosla13cvpr}
use web images as a prior to summarize user generated videos. Each
of these methods use a different technique for evaluating the quality
of their video summarization approach. In order to address this issue,
our focus in this paper is to introduce an evaluation technique that
can automatically evaluate the quality of video summaries.

\textbf{Techniques for evaluating video summaries}: Most previous
work evaluate the performance of their video summarization techniques
using user studies \cite{Ngo01jig,Doherty08multimedia,Lee12cvpr,Lu13cvpr}.
User study requires re-comparison every time algorithm parameters
are tweaked and is difficult for others to replicate. Liu et al. \cite{Liu10pami}
measure the performance based on the presence of objects of interest.
Li and Maerialdo \cite{Li10cbmi} and Khosla et al. \cite{Khosla13cvpr}
use pixel-based distance of a summary to the original video for evaluation.
The drawback of using pixel-based distance is that it does not necessarily
measure the semantic similarity between subshots, but rather forces
them to be similar in color and texture space. Li and Maerialdo \cite{Li10mm}
introduce VERT, which evaluates video summaries given a ground-truth
video summary by counting the number of sub-shots that overlap between
the two. This method also suffers from the disadvantage of pixel-based
distance. In addition, people often find it a hard task to generate
a ground-truth video summary, whereas they are more comfortable summarizing
video in text. In constrast to these techniques, we introduce a method
that transfers the video summary evaluation problem into the text
domain and measures the semantic similarity between automatically
generated summaries and ground-truth summaries.

\textbf{Techniques for evaluating text summaries}: In constrast to
the field of computer vision, there has been large progress in the
NLP community on evaluating text summaries. The first techniques in
NLP were created in order to evaluate the quality of text which had
been machine translated from one language to another \cite{Doddington02hlt,Papineni02acl}.
Later on, Lin \cite{Lin04was} introduced ROUGE for evaluating video
summaries. The algorithms in ROUGE are inspired by the methods for
evaluating machine translation. There have been other more recent
techniques for evaluating text summaries \cite{hovy2006automated,zhou2006paraeval,giannakopoulos2008summarization},
but ROUGE still remains the standard evaluation algorithm. In this
paper, we map the video summary evaluation problem into the text domain
and use ROUGE to measure the similarity between the summaries.

\section{Evaluation Framework}

In Sec. 3.1, we provide an overview of VideoSET, and describe how
it can be used to evaluate video summaries. Then in Sec. 3.2, we describe
the video datasets for which we have obtained text annotations and
ground-truth summaries that can be used in VideoSET. Finally, in Secs.
3.2-3.5, we explain each component of the framework in detail: obtaining
text annotations, obtaining ground-truth summaries, generating a text
representation of a video summary, and scoring the video summary.

\begin{figure}
\begin{centering}
\includegraphics[scale=0.6]{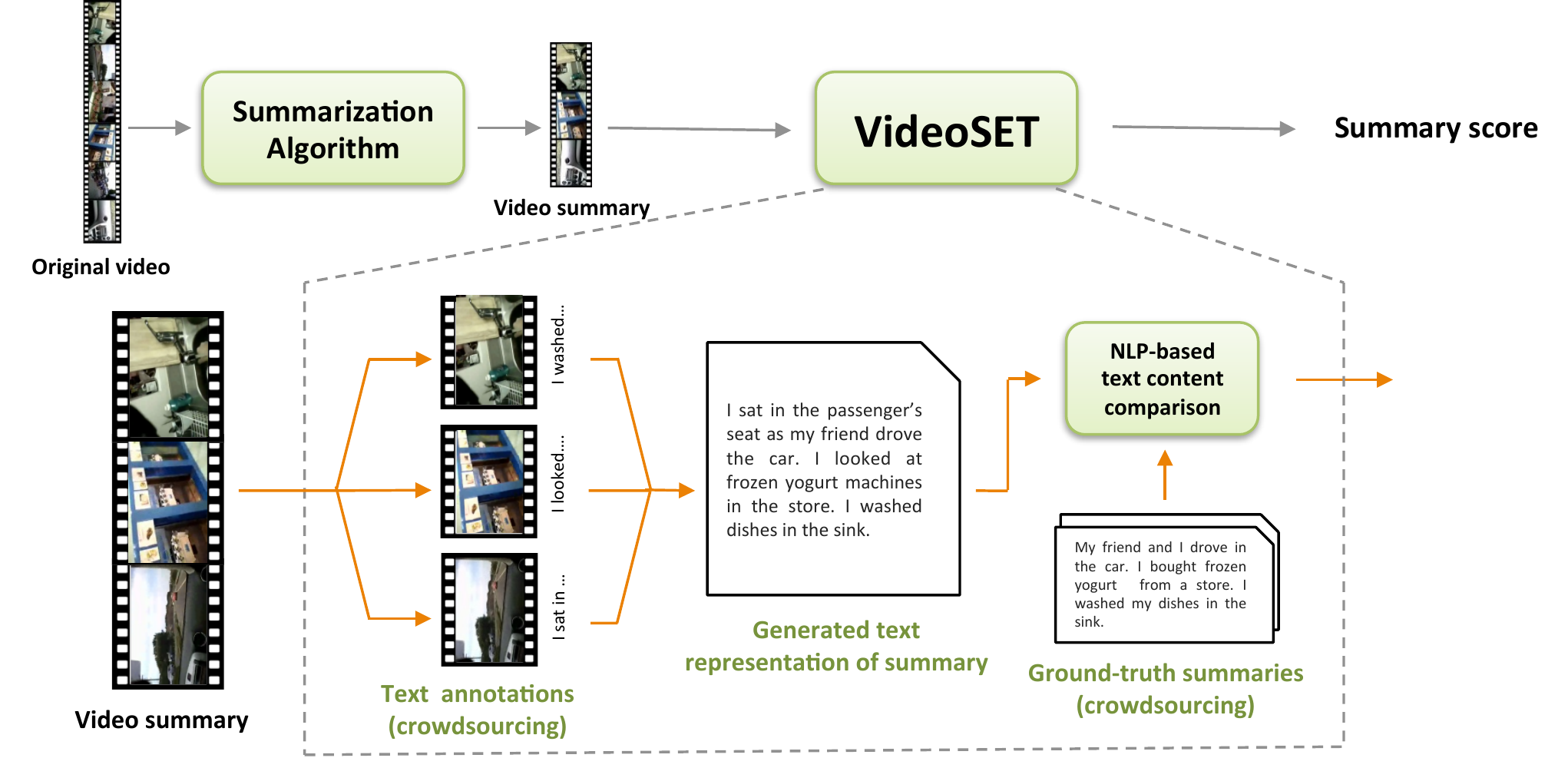}
\par\end{centering}

\caption{Overview of VideoSET evaluation framework. Given a video summary,
a text representation is generated using text annotations from the
original video. The text representation is then compared against human-written
ground-truth summaries, using NLP-based text content comparison. The
summary score is returned as output. Text annotations and ground-truth
summaries are released for a number of video datasets.\label{fig:Illustration}}
\end{figure}

\subsection{Overview of VideoSET}

Fig. \ref{fig:Illustration} provides an overview of VideoSET. A video
is represented as a sequence of $M$ subshots $\mathcal{V}=\{v_i\}_{i=1}^{M}$,
and a video summary is a subset $\mathcal{C} \subset \mathcal{V}$
of these subshots. A user constructs a video summary using a summarization
algorithm, and provides it to VideoSET as input. VideoSET then generates
a text representation $T(\mathcal{C})$ of the summary, using text
annotations of the original video. The text representation is compared
against a set of ground-truth text summaries $\mathcal{G}$, that
are written by humans to specify the ideal semantic content of a video
summary. We have released all necessary text annotations and ground-truth
summaries for a number of video datasets.

Comparison against the ground-truth text summaries is performed using
a scoring function

\begin{equation}f(\mathcal{C},\mathcal{G}) = \max_{g_i \in \mathcal{G}} S(T(\mathcal{C}),g_i)\end{equation}where
$S(x,y)$ is a function that measures the semantic similarity of texts
$x$ and $y$. For $S(x,y)$ we use the ROUGE metric that is a standard
for text summary evaluation. The evaluation score is then returned
to the user as output.

\subsection{Datasets\label{sub:Datasets}}

\begin{figure}
\begin{centering}
\includegraphics[scale=0.65]{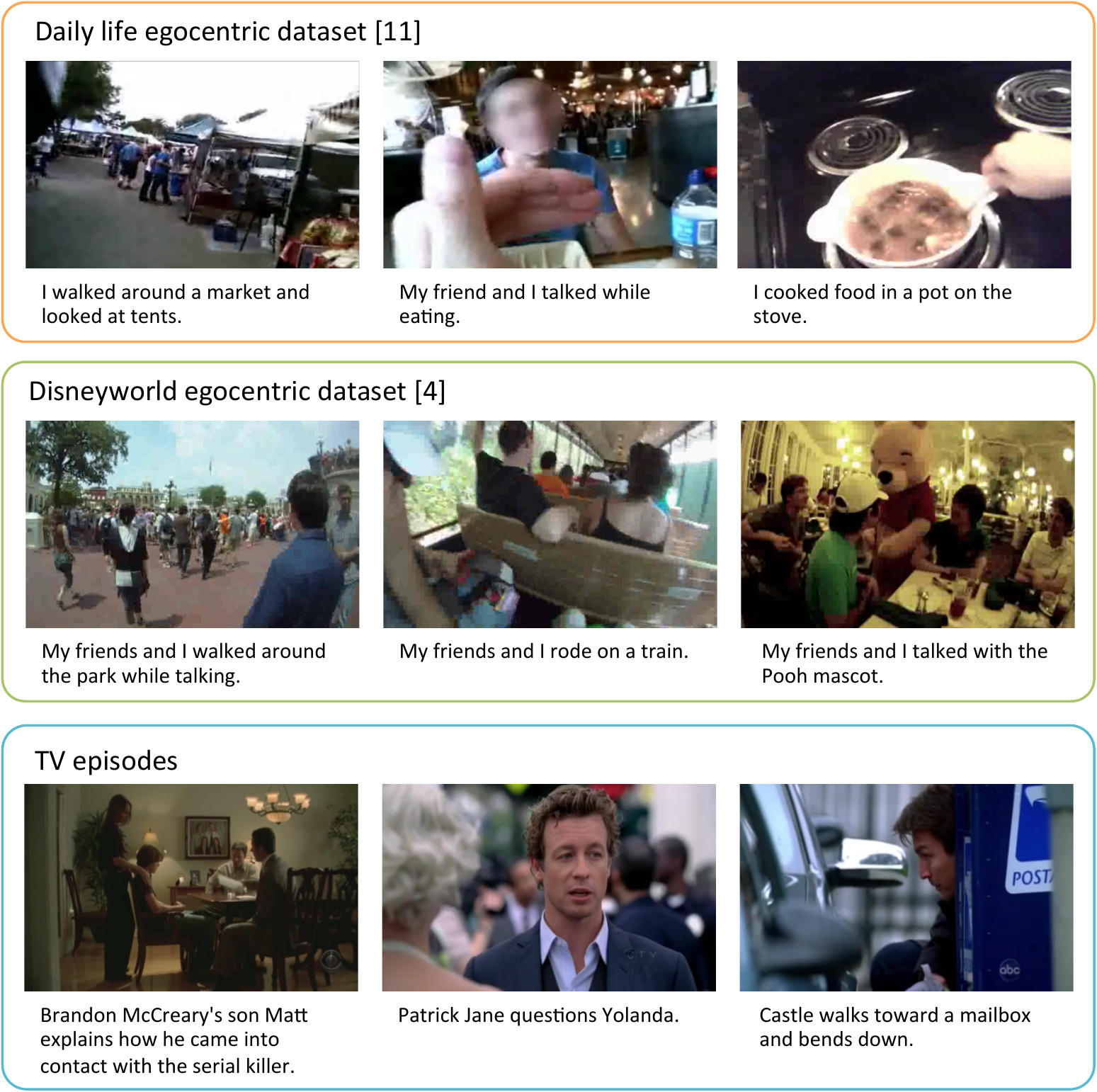}
\par\end{centering}

\caption{Datasets for which VideoSET provides all required text annotations
and ground-truth summaries.\label{fig:Datasets}}
\end{figure}

We have released text annotations and ground-truth summaries that
can be used in VideoSET for two publicly available egocentric video
datasets, and four TV episodes. Each of these are described in more
detail below, and representative images and text annotations are shown
in Fig. \ref{fig:Datasets}.

\textbf{Daily life egocentric dataset} \cite{Lee12cvpr}\textbf{ }This
dataset consists of 4 egocentric videos of 3-5 hours each. Each video
records a subject through natural daily activities such as eating,
shopping, and cooking. The videos were recorded using a Looxcie wearable
camera at 15 fps and 320$\times$480 resolution. We provide text annotations
and ground-truth summaries for all videos in this dataset.

\textbf{Disneyworld egocentric dataset} \cite{Fathi12cvpr} This dataset
consists of 8 egocentric videos of 6-8 hours each. Each video records
a subject during a day at Disneyworld Park. The videos were recorded
using a GoPro wearable camera at 30 fps and 1280 $\times$ 720 resolution.
We provide text annotations and ground-truth summaries for 3 videos
in this dataset.

\textbf{TV episodes} We provide text annotations and ground-truth
summaries for 4 TV episodes of 45 minutes each. The episodes consist
of 1 from Castle, 1 from The Mentalist, and 2 from Numb3rs.

In all, we provide annotations for 40 hours of data split over 11
videos. Our annotations may also be of interest to researchers working
in the intersection between images or video and text, similar to \cite{ordonez2011im2text}
and \cite{guadarramayoutube2text}.

\subsection{Obtaining text annotations\label{sub:Obtaining-text-annotations}}

We segmented egocentric videos from the datasets in Sec. \ref{sub:Datasets}
into 5-second subshots, and TV episodes into 10-second subshots. We
then obtained 1-sentence descriptions of each subshot using Amazon's
Mechanical Turk. Workers were asked to write a simple and factual
sentence about what happened in each subshot. They were instructed
to write from a first-person past-tense perspective for the egocentric
videos, and from a third-person present-tense perspective for the
TV episodes. Workers who annotated the TV episodes were required to
be familiar with the episode, and to use the TV character names in
their descriptions. The descriptions were edited by additional workers
for vocabulary and grammatical consistency.

\subsubsection*{Choosing subshot length}

To choose the subshot length, we first obtained text annotations for
an egocentric video at 3, 5, and 10 seconds, and for a TV episode
at 5, 10, and 20 seconds. The shortest subshot length for each type
of video was chosen to be sufficiently fine to oversegment the video.
We then used the ROUGE content similarity metric to compute the similarity
between the text annotations at each subshot length. The similarity
across the different subshot lengths was high, indicating that content
coverage was preserved across the different lengths. Any of the lengths
would be appropriate using our framework. We therefore chose to use
5-second subshots for the egocentric videos and 10-second subshots
for the TV episodes, to balance the trade-off between having as fine-grained
annotations as possible and minimizing the cost of obtaining the annotations.

While VideoSET is designed to evaluate summaries in the form of specific-length
subshots, it can easily be adapted and used to evaluate summaries
in other formats as well. For example, a summary consisting of keyframes
can be represented in text using the annotations for the subshot containing
each keyframe. This is appropriate since our subshots are short enough
to express a single semantic concept or event. A summary consisting
of variable-length subshots can also be evaluated by mapping the subshots
to appropriate text annotations.

\subsection{Obtaining ground-truth summaries\label{sub:Obtaining-ground-truth-summaries}}

\begin{figure}
\begin{centering}
\includegraphics[scale=0.6]{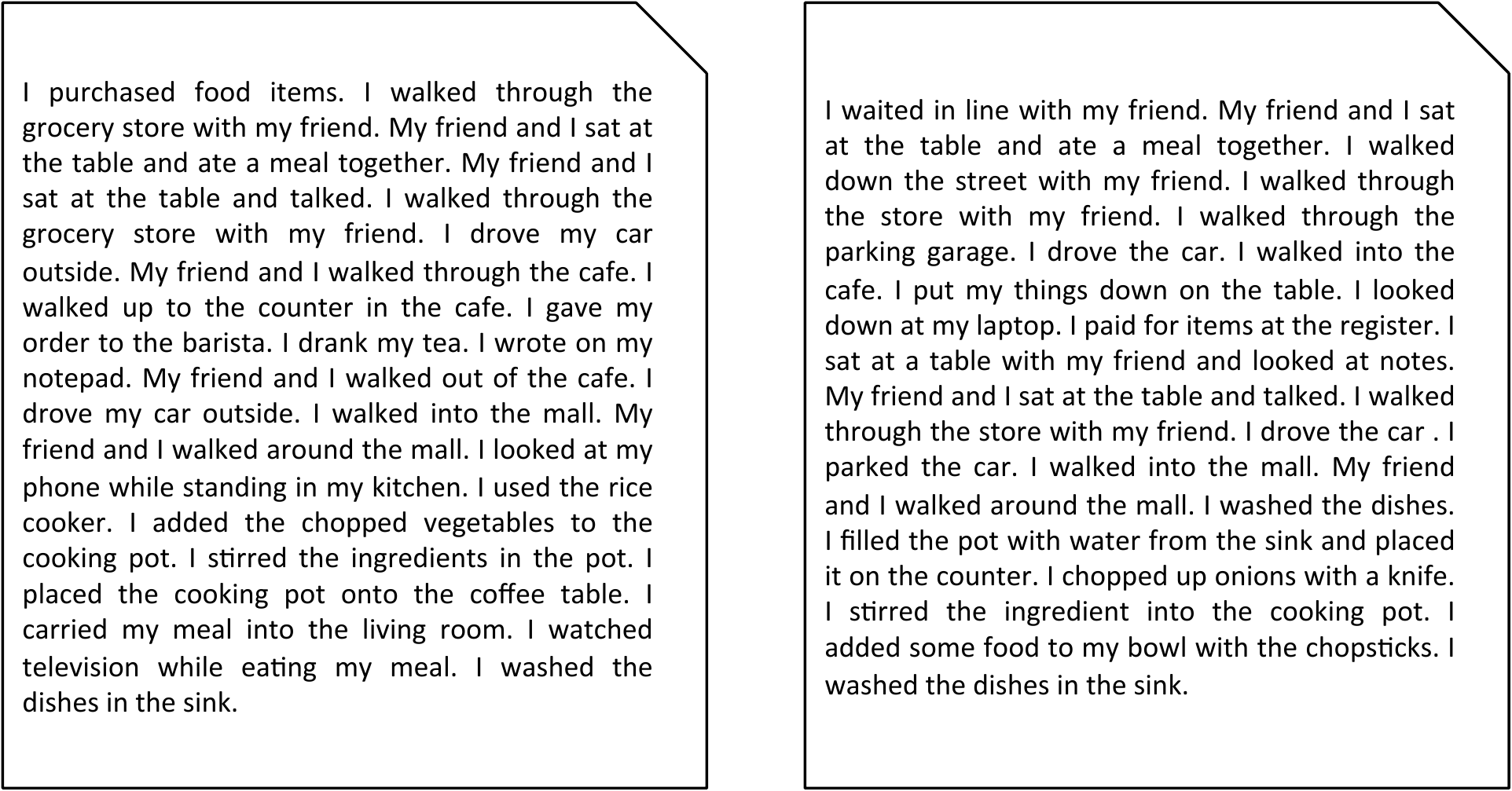}
\par\end{centering}

\caption{Example of two different ground-truth summaries for a video.\label{fig:sample-reference}}
\end{figure}

We obtained ground-truth summaries for videos in text form, since
humans can most naturally express semantic information through words.
It is also easier for humans to write down the information they feel
should be in a summary, than it is to comb through a long video and
pick out the ideal subshots. For example, it may be clear that a summary
should show that the camera-wearer ``walked on the sidewalk.'' However,
as the examples in Fig. \ref{fig:Pixel-based-comparison-does} show,
many visually diverse and equally good subshots can illustrate this
and it is unclear which should be included in a ground-truth.

We asked a small group of workers to write a summary in words about
what happened in each video. The workers were provided with the text
annotations for the video so that similar vocabulary could be used.
They were asked to write simple sentences with a similar level of
content as the text annotations. They were also asked to rank their
sentences in order of importance. Then during the evaluation process,
a video summary of $|\mathcal{C}|$ subshots is compared with a length-adjusted
ground-truth summary consisting of the top $|\mathcal{C}|$ most important
sentences in temporal order. 

Fig. \ref{fig:sample-reference} shows an example of length-adjusted,
24-sentence ground-truth summaries written by two different workers.
Workers typically wrote and ranked between 40-60 summary sentences
per egocentric video, and 20-30 sentences per TV episode.

\subsection{Generating the text representation of a video summary\label{sub:Generating-a-text-1}}

\begin{figure*}
\begin{centering}
\includegraphics[scale=0.27]{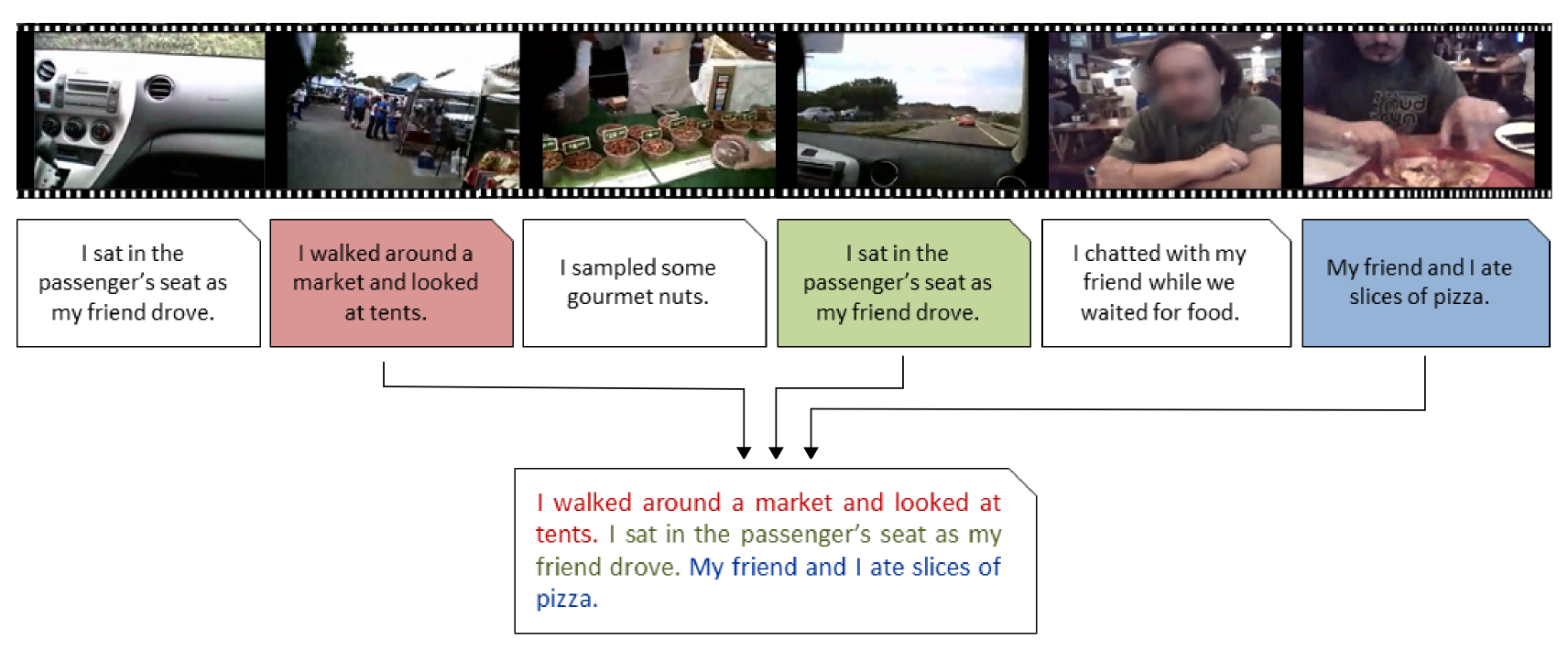}
\par\end{centering}

\caption{The text representation of a video summary is generated by concatenating
text annotations for the subshots in the summary.\label{fig:A-text-representation}}
\end{figure*}

Given a video summary $\mathcal{C}$ to evaluate, VideoSET first generates
a text representation $T(\mathcal{C})$ of the summary. This representation
can be acquired by concatenating the pre-existing text annotations
(Sec. \ref{sub:Obtaining-text-annotations}) associated with each
summary subshot, since the summary is a collection of subshots from
the original video. We have released text annotations for the videos in Sec. \ref{sub:Datasets} so
that no effort is required on the part of the user, and the process
is illustrated in Fig. \ref{fig:A-text-representation}.

\subsection{Scoring the video summary}

To score the video summary, a similarity function $S(x,y)$ is used
to compare the text representation of the summary with ground-truth
text summaries. We use the ROUGE-SU metric from the publicly available
ROUGE toolbox \cite{Lin04was}. ROUGE-SU measures unigram and skip-bigram
co-occurence between a candidate and ground-truth summary, after pre-processing
to stem words and remove stopwords. Skip-bigrams are any pair of words
in their sentence order, allowing for arbitrary gaps. For example,
the skip-bigrams for the sentence ``I walked my dog at the park.''
are: ``walked dog'', ``walked park'', and ``dog park'', where
stopwords have not been included. The unigrams and skip-bigrams are
treated equally as counting units. We use the F-measure for ROUGE-SU.

The ROUGE toolbox is a collection of n-gram comparison metrics that
measure text content similarity, and more detail can be found in \cite{Lin04was}.
We ran experiments using each of the metrics in ROUGE and found ROUGE-SU
to have the strongest correlation with human judgment.

\subsubsection*{Addressing human subjectivity\label{sub:Addressing-human-subjectivity}}

To address human subjectivity about what is semantically most important,
we use the approach of ROUGE to compare a video summary with multiple
ground-truth summaries. The score of a video summary $\mathcal{C}$
with respect to a set of ground-truth summaries $\mathcal{G}$ is
computed as $f(\mathcal{C},\mathcal{G}) = \max_{g_i \in \mathcal{C}} S(T(\mathcal{G}),g_i)$,
the maximum of pairwise summary-level scores between the video summary
and each ground-truth. We have released 3 ground-truth summaries for
each video in in Sec. \ref{sub:Datasets}, and since writing a ground-truth summary is a quick and
simple task, this number can be easily scaled in the future.

\section{Experiments\label{sec:Experiments}}

To assess the effectiveness of VideoSET, we conducted two different
experiments. In the first experiment, we generated a number of video
summaries using existing video summarization methods, and correlated
their VideoSET scores with human judgment. In the second experiment,
we analyzed VideoSET's performance in the full space of possible video
summaries. We randomly sampled pairs of video summaries and
subshots, and compared VideoSET judgment with human judgment.

To confirm our intuition that text distance is more appropriate than
visual distance as a measure of semantic similarity, we also compare
with a pixel-based distance metric in our experiments.

\subsection{VideoSET evaluation of existing summarization methods \label{sub:baselines}}

We generated video summaries using the following existing summarization
methods. 2-minute summaries ($N=24$ subshots for egocentric video
and $N=12$ subshots for TV episodes) were generated using each method.
\begin{enumerate}
\item \textbf{Uniform sampling: }$N$ subshots uniformly spaced throughout
the original video were selected.
\item \textbf{Color histogram clustering:} Frames extracted at 1fps were
clustered into $N$ clusters using $\chi^2$-distance between color
histograms of the frames. Subshots containing the frame closest to
the center of each of the $N$ clusters were selected for the video
summary.
\item \textbf{Video-MMR }\cite{Li10cbmi}\textbf{:} Frames were extracted at 1fps from the original
video. In each of $N$ iterations, a keyframe was chosen that was
most visually similar to the frames not yet selected as keyframes,
and at the same time different from the frames already selected as
keyframes. In other words, each iteratively selected keyframe has
Maximal Marginal Relevance (MMR). Given the set of all video frames
$V$ and the set of already-selected keyframes $S_{n-1} = \{s_1,...,s_{n-1}\}$,
the $n$th keyframe $s_n$ is selected as \begin{equation}s_n = \arg\min_{f \in V\backslash S_{n-1}}(\lambda \chi^2(f,V\backslash S_{n-1}) - (1-\lambda)\min_{s \in S_{n-1}} \chi^2(f,s)) \end{equation}
$\lambda$ was empirically chosen to be 0.5. Subshots containing the
chosen keyframes were selected for the video summary.
\item \textbf{Object-driven summarization }\cite{Lee12cvpr}\textbf{:}
The method of Lee et al. \cite{Lee12cvpr} chooses keyframes containing
important people and objects based on a learned metric for importance.
Keyframe summaries were provided by the authors for the videos in
the Daily life egocentric dataset. The subshots containing the keyframes
were selected for the video summary. 
\end{enumerate}
We also generated summaries using two additional methods that utilize
the ground-truth text summaries and text annotations. These methods
attempt to maximize our metric score given the ground-truth summaries
and thus represent summaries close to what our metric would consider
ideal.
\begin{enumerate}
\item \textbf{Greedy BOW:} The words in the ground-truth summary were considered
as an unordered ``bag of words.'' Subshots were greedily selected
based on unigram matching of the subshots' text annotations with the
ground-truth bag of words.
\item \textbf{Sentence-based Ordered Subshot Selection:} One subshot was
selected for each sentence in the ground-truth summary, using a dynamic
programming approach that restricted the selected subshots to be in
the same relative order as the corresponding sentences.
\end{enumerate}
We computed VideoSET scores for video summaries generated using the
above methods, for all the videos in the datasets of Sec. \ref{sub:Datasets}.
For a summary length of 2 minutes, 24 video summaries were generated
for the Egocentric daily life dataset (6 methods x 4 original videos),
15 video summaries were generated for the Disney egocentric dataset
(5 methods x 3 original videos), and 20 video summaries were generated
for the TV episodes (5 methods x 4 episodes). We also computed scores
for each of these videos using a pixel-based distance metric for comparison.
The pixel-based distance metric was defined as the average visual
similarity of the summary subshots to human-defined ground-truth summary
subshots, based on minimum $\chi^2$-color histogram distance of the
frames in a subshot to the ground-truth subshot frames.

We correlated rankings based on the VideoSET and pixel-based scores
with human rankings from a user study. Humans were asked to rank the
video summaries generated using the above methods, in terms of
how semantically similar they were to the content of ground-truth
written summaries. The score was taken to be the highest score with
respect to 3 ground-truth summaries. The Spearman's rank order correlation
coefficient between each of the automated metrics and the human-assigned
ranks from this study are shown in Table \ref{tab:Spearman-correlation-coefficient}.

The results in Table \ref{tab:Spearman-correlation-coefficient} show
that VideoSET is strongly correlated with human judgment, and has
better performance than a pixel-based distance metric. The largest
correlation gap between VideoSET and the pixel-based distance is for
the Disney dataset, which is most challenging due to the highly varied
visual scenes as the subjects tour through the amusement park. The
smallest correlation gap is for the TV episodes, where both methods
perform strongly due to the fact that TV shows are highly edited with
little redundancy.

\begin{table*}
\begin{centering}
\begin{tabular}{|>{\centering}p{0.25\textwidth}|>{\centering}m{0.25\textwidth}|>{\centering}m{0.23\textwidth}|>{\centering}m{0.23\textwidth}|}
\hline 
 & Daily life dataset \cite{Lee12cvpr} & Disney dataset \cite{Fathi12cvpr} & TV episodes\tabularnewline
\hline 
\hline 
VideoSET & \textbf{0.83} & \textbf{0.96} & \textbf{0.98}\tabularnewline
\hline 
Pixel-based distance & 0.73 & 0.48 & 0.93\tabularnewline
\hline 
\end{tabular}
\par\end{centering}

\caption{Spearman correlation coefficients of human judgment with automated
evaluation of video summaries generated using existing video summarization
methods.\label{tab:Spearman-correlation-coefficient}}
\end{table*}

\subsection{VideoSET Evaluation of Randomly Sampled Summaries and Subshots\label{sub:Comparing-visual-versus}}

To better understand VideoSET's performance in the full space of possible
summaries, we randomly sampled video summaries as well as subshots,
and compared VideoSET judgment with human judgment. 

We first randomly generated 100 pairs of 2-min. summaries (24 subshots)
for a video in the Daily life egocentric dataset \cite{Lee12cvpr}.
We asked two humans to watch each pair of summaries and judge which
was semantically closer to a provided ground-truth text summary. In
40\% of the comparisons, the two human judges disagreed, indicating
that the difference was too ambiguous even for humans. For the remaining
60\% of the comparisons, we computed automated judgments using VideoSET
scores as well as a pixel-based distance metric. The results are shown
in Table \ref{tab:sample-summaries}, and show that VideoSET scores
have higher agreement with human judgment than the pixel-based distance
metric.

\begin{table}
\begin{centering}
\begin{tabular}{|>{\centering}p{0.4\textwidth}|>{\centering}p{0.2\textwidth}|>{\centering}p{0.2\textwidth}|}
\hline 
 & VideoSET & Pixel-based\tabularnewline
\hline 
\hline 
Agreement with humans (\%) & \textbf{61.0} & 52.5\tabularnewline
\hline 
\end{tabular}
\par\end{centering}

\caption{Agreement of VideoSET and pixel-based distance with human judgment,
when choosing which of a pair of randomly generated 2-min summaries
is semantically closer to a provided ground-truth text summary. 100 pairs
of summaries were evaluated.\label{tab:sample-summaries}}
\end{table}

At a finer level, we then assessed the performance of VideoSET on
comparing pairs of individual subshots. Since the space is now more
constrained, we densely computed VideoSET scores for every pair of
subshots in the video with respect to every possible third subshot
as a reference. We also computed scores based on the pixel-based distance
metric. Based on these, we separated the comparisons into 4 different
cases: (1) VideoSET judged both subshots to have no semantic similarity
with the reference subshot; (2) VideoSET judged both subshots to have
equal, non-zero semantic similarity with the reference subshot; (3)
VideoSET judged one subshot to be semantically more similar than the
other, and \emph{agreed} with the pixel-based (PB) judgment; and (4)
VideoSET judged one subshot to be semantically more similar than the
other, and \emph{disagreed} with the pixel-based (PB) judgment. We
then sampled 300 comparisons from each of these 4 cases (a total of
1200 comparisons). For these samples, we asked humans to judge which subshot in each
pair is semantically more similar to the reference subshot, if the
pair is equally similar, or if both subshots have no similarity. The
agreement of the VideoSET and pixel-based judgments with the human
judgments is shown in Table \ref{tab:sample-subshots}. 

\begin{table}
\begin{centering}
\begin{tabular}{|>{\centering}p{0.18\textwidth}|>{\centering}p{0.08\textwidth}|>{\centering}p{0.15\textwidth}|>{\centering}p{0.15\textwidth}|>{\centering}p{0.18\textwidth}|>{\centering}p{0.18\textwidth}|}
\cline{3-6} 
\multicolumn{2}{c|}{} & \multicolumn{2}{>{\centering}p{0.3\textwidth}|}{{\scriptsize \% Correct}{\scriptsize \par}

{\scriptsize (Using all human judgments)}} & \multicolumn{2}{>{\centering}p{0.36\textwidth}|}{{\scriptsize \% Correct}{\scriptsize \par}

{\scriptsize (Using non-zero human judgments)}}\tabularnewline
\hline 
{\scriptsize VideoSET judgment } & {\scriptsize \% of cases} & {\scriptsize VideoSET} & {\scriptsize Pixel-based} & {\scriptsize VideoSET} & {\scriptsize Pixel-based}\tabularnewline
\hline 
\hline 
{\scriptsize Both zero } & {\scriptsize 65.1} & \textbf{\scriptsize 91.0} & --- & \textbf{\scriptsize ---} & {\scriptsize 54.4}\tabularnewline
\hline 
{\scriptsize Both equal } & {\scriptsize 2.4} & \textbf{\scriptsize 29.0} & {\scriptsize 20.3} & \textbf{\scriptsize 52.7} & {\scriptsize 36.9}\tabularnewline
\hline 
{\scriptsize Inequal, }\emph{\scriptsize agrees}{\scriptsize{} with
PB } & {\scriptsize 22.5} & \textbf{\scriptsize 48.5} & {\scriptsize 48.5} & \textbf{\scriptsize 91.5} & {\scriptsize 91.5}\tabularnewline
\hline 
{\scriptsize Inequal, }\emph{\scriptsize disagrees}{\scriptsize{} with
PB} & {\scriptsize 10.0} & \textbf{\scriptsize 18.2} & {\scriptsize 8.8} & \textbf{\scriptsize 53.6} & {\scriptsize 25.8}\tabularnewline
\hline 
\end{tabular}
\par\end{centering}

\caption{Agreement of VideoSET with human judgment, when choosing which of
a pair of different subshots is semantically closest to a reference
subshot. The comparisons are separated according to the VideoSET judgment, and the \% of all cases for which the judgment occurs is listed. For
each type of judgment, the \% correct of VideoSET with respect to
human judgment for 300 sampled comparisons is given, as well as the
\% correct of a pixel-based distance metric. Agreement using only
non-zero human judgments in addition to all human judgments is given,
since the large majority of human judgments evaluate both subshots
in a pair to have zero similarity with the reference subshot. PB stands
for pixel-based distance metric. \label{tab:sample-subshots} }
\end{table}

\begin{figure}
\begin{centering}
\includegraphics[scale=0.9]{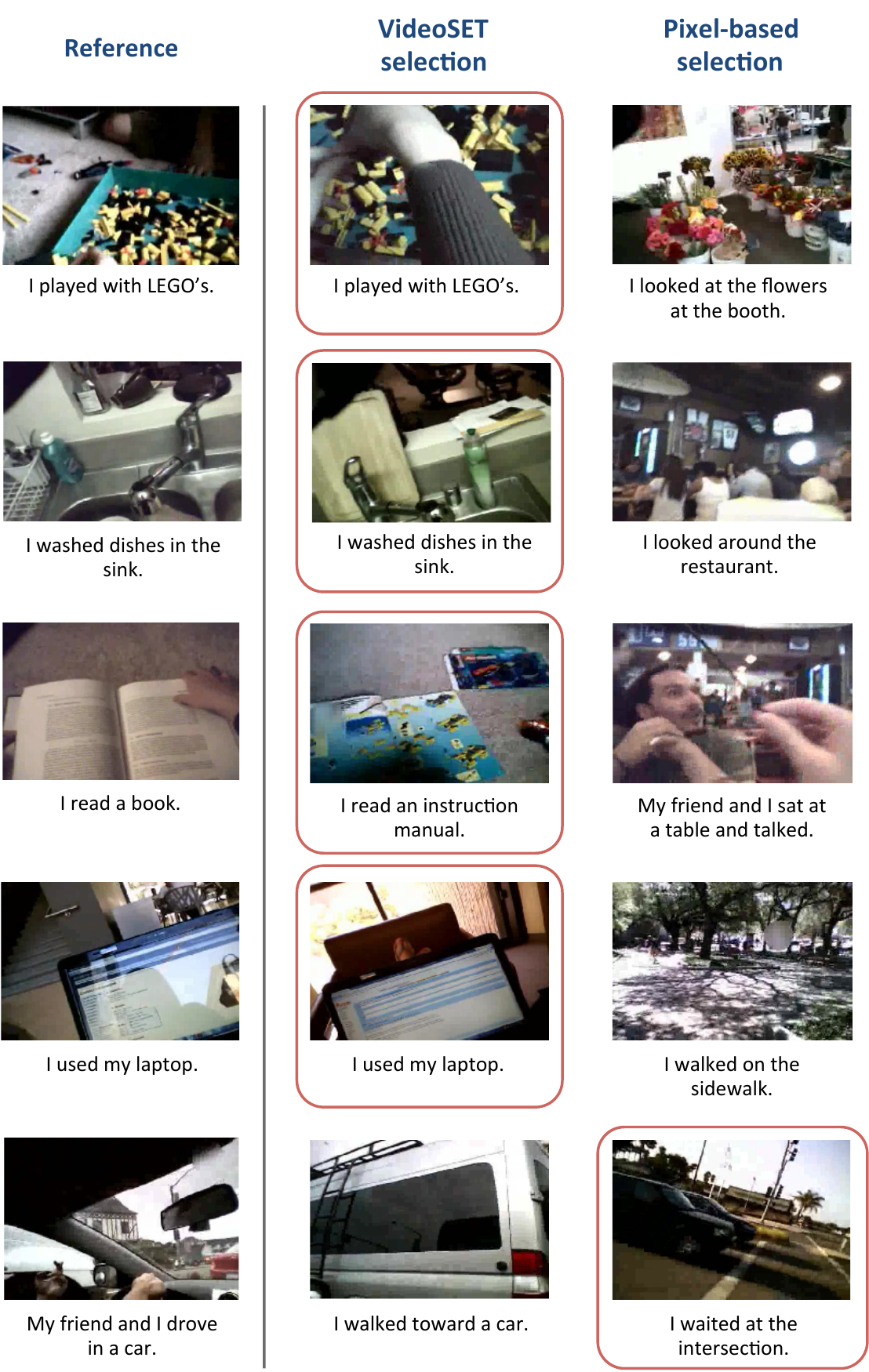}
\par\end{centering}

\caption{Examples of comparisons where VideoSET and the pixel-based metric
disagree. Each row shows a representative image from the reference
subshot, the subshot that VideoSET selected, and the subshot that
the pixel-based metric selected. The human selection for each example
is boxed. VideoSET and the pixel-based metric disagree in all the
examples shown.\label{fig:Examples-from-the}}
\end{figure}

Both VideoSET and humans judged the majority of subshots to have zero
similarity (65.1\% of comparisons for VideoSET, 77.3\% for humans).
This is expected since most pairs of subshots should not be semantically
related. Because of this, we also show the agreements using only non-zero
human judgments. The results indicate that VideoSET has stronger agreement
with human judgment than the pixel-based metric. Additionally, when
VideoSET and the pixel-based metric both judge that one subshot is
semantically closer than the other but disagree, VideoSET agrees with
human judgment more than twice as often as the pixel-based metric.
Some illustrative examples of comparisons where VideoSET and the pixel-based
metric disagree are shown in Fig. \ref{fig:Examples-from-the}.

\section{Conclusion}

We have developed an evaluation technique to automatically measure
how well a video summary retains the semantic information in the original
video. Our approach is based on generating a text representation of
the video summary, and measuring the semantic distance of the text
to ground-truth text summaries written by humans. Our experiments
show that this approach correlates well with human judgment, and
outperforms pixel-based distance measures. In addition, our framework
can be extended to evaluate any type of video summary, and can accommodate
future extensions to our semantic distance metric.

\section{Acknowledgements}
This research is partially supported by an ONR MURI grant and an Intel gift, and a Stanford Graduate Fellowship to S.Y.

\bibliographystyle{plain}
\bibliography{refs}

\end{document}